\documentclass{article}
\usepackage{ijcai24}

\usepackage{times}
\usepackage{soul}
\usepackage{url}
\usepackage[hidelinks]{hyperref}
\usepackage[utf8]{inputenc}
\usepackage[small]{caption}
\usepackage{graphicx}
\usepackage{amsmath}
\usepackage{amsthm}
\usepackage{booktabs}
\usepackage{algorithm}
\usepackage{algorithmic}
\usepackage[switch]{lineno}

\usepackage{natbib}
\usepackage{subcaption}
\usepackage{graphicx}

\usepackage{xcolor, colortbl}
\definecolor{lightcream}{RGB}{255,244,212}
\sethlcolor{lightcream}
\usepackage{fancyhdr}
\usepackage{hyperref}

\title{Dilated Convolution with Learnable Spacings makes visual models more aligned with humans: a Grad-CAM study}

\author{
Rabih Chamas$^1$
\and
Ismail Khalfaoui-Hassani$^3$\and
Timothée Masquelier$^{2,3}$\\
\affiliations
$^1$LIS, CNRS.\\
$^2$CerCo UMR 5549, CNRS. 
$^3$Université de Toulouse, France.\\
\emails
rabih.chamas@lis-lab.fr,
ismail.khalfaoui-hassani@univ-tlse3.fr,
timothee.masquelier@cnrs.fr
}

\begin{document}
\maketitle

\begin{abstract}
Dilated Convolution with Learnable Spacing (DCLS) is a recent advanced convolution method that allows enlarging the receptive fields (RF) without increasing the number of parameters, like the dilated convolution, yet without imposing a regular grid. DCLS has been shown to outperform the standard and dilated convolutions on several computer vision benchmarks. Here, we show that, in addition, DCLS increases the models' interpretability, defined as the alignment with human visual strategies. To quantify it, we use the Spearman correlation between the models’ Grad-CAM heatmaps and the ClickMe dataset heatmaps, which reflect human visual attention. We took eight reference models – ResNet50, ConvNeXt (T, S and B), CAFormer, ConvFormer, and FastViT (sa\_24 and 36) – and drop-in replaced the standard convolution layers with DCLS ones. This improved the interpretability score in seven of them. Moreover, we observed that Grad-CAM generated random heatmaps for two models in our study: CAFormer and ConvFormer models, leading to low interpretability scores. We addressed this issue by introducing Threshold-Grad-CAM, a modification built on top of Grad-CAM that enhanced interpretability across nearly all models. The code and checkpoints to reproduce this study are available at: \href{https://github.com/rabihchamas/DCLS-GradCAM-Eval}{https://github.com/rabihchamas/DCLS-GradCAM-Eval}
\end{abstract}

\section{Introduction}
\label{sec:intro}

Deep learning neural networks are extremely powerful for a myriad of tasks, including image classification. However, despite being very powerful tools, they remain black box models, and understanding how they arrive at their results can be a major challenge. Explainability methods in deep learning aim to explain why a particular model predicts a particular result.

One application where explainability methods have been successfully in use for several years is image classification. The most popular and successful models nowadays for image classification include convolution and/or attention layers. 

When a model contains only convolutions, it is called a fully convolutional neural network or CNN, when it contains only multi-head self-attention (MHSA) layers, it is called a transformer or, in the context of computer vision, a vision transformer, and when a model contains both layers, it is called a hybrid model.

Despite their very high accuracy, most of these models remain very opaque, and the lack of explicability of the latter, especially in computer vision, raises concerns about trust, fairness, and interoperability, hindering their adoption in sensitive areas such as medical diagnosis \citep{COLLENNE2024} or autonomous vehicles.

This also applies to recent advances such as Dilated Convolution with Learnable Spacings (DCLS) \citep{hassani2023dilated}, which shows promising performance gains in tasks such as image classification, segmentation, and object detection.

While the accuracy of DCLS is encouraging, its black-box nature demands attention. Thus, we have been motivated to explore explainability measures and scores specifically for DCLS, with the hope of shedding light on its underlying decision-making processes.

The taxonomies used for explainability in the artificial intelligence field of research are diverse and constantly evolving as new approaches are discovered. However, a common way to proceed is to distinguish between two major families of model explainability methods: global methods and local methods \citep{speith2022review,schwalbe2023comprehensive}.

Global methods describe the overall behavior of the model, considering general patterns and the importance of features. Some examples include Partial Dependence Plots (PDPs) \citep{friedman2001greedy} and SHapley Additive explanations (SHAP) \citep{lundberg2017unified}. Local methods, on the other hand, focus on explaining individual predictions, focusing on why the model made a particular decision for a particular input. Examples include Local Interpretable Model-Agnostic Explanations (LIME) \citep{ribeiro2016should} and Gradient-weighted Class Activation Mapping (Grad-CAM) \citep{selvaraju2017grad}.

 Grad-CAM is a popular method that helps visualize which parts of an image are most important to the model's decision. For the needs of our study, we designed a new explainability method based on Grad-CAM that we called Thershold-Grad-CAM. This new explainability method overcomes some issues tied to the failure of traditional Grad-CAM, in particular, for ConvFormer and CAFormer architectures \citep{yu2023metaformer}.

The objective of this paper is, on the one hand, to perform a comparative study in terms of explainability scores between recent state-of-the-art models in computer vision, namely ConvNeXt \citep{liu2022convnet}, ConvFormer \citep{yu2023metaformer}, CAFormer \citep{yu2023metaformer}, and FastViT \citep{vasu2023fastvit} in their original form, and, on the other hand, to perform the same study between these same models and their DCLS-enhanced counterparts. 

\begin{figure*}[!htbp]
  \centering

    \includegraphics[width=\linewidth]{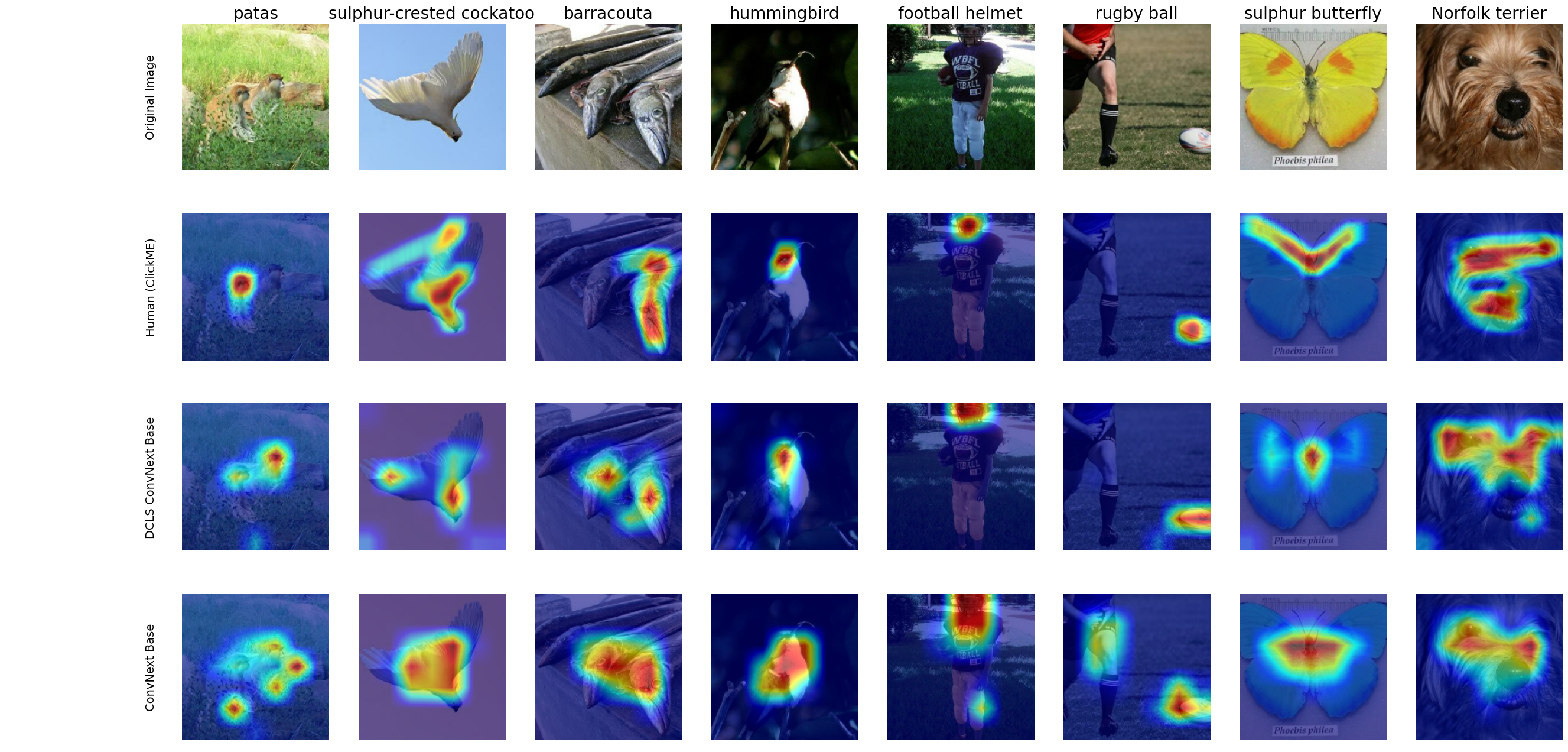} % 
    \caption{Visualization of Heatmaps on ClickMe dataset Images. First row: original images from the ClickMe dataset. Second row: the same images superimposed with heatmaps created by humans from the ClickMe project. Third row: Threshold-GradCAM heatmaps of the ConvNeXt base model enhanced with DCLS. Fourth row: Threshold-GradCAM heatmaps of the baseline ConvNeXt base model without DCLS.}
    \label{fig:qualitatif_results}
\end{figure*}

What motivated the comparative study presented here in this paper is the qualitative similarity we noticed between human attention heatmaps obtained from the ClickMe dataset \citep{linsley2019learning} and those obtained by models empowered with DCLS. Figure~\ref{fig:qualitatif_results} gives an overview of this similarity based on the ConvNeXt-B model. The images presented in the figure~\ref{fig:qualitatif_results} were selected from the ClickMe dataset. To help illustrate our point, we have selected a few images where the heatmaps are visually relevant. We then quantitatively confirmed this remarkable alignment of DCLS models and human attention heatmaps through a rigorous study of the Spearman correlation \citep{zar2005spearman} between heatmaps generated by Threshold-Grad-CAM and heatmaps made by human participants in the ClickMe dataset.%, which provides insights into human attention patterns during object recognition.

We will refer to a model empowered with DCLS by its original name followed by the suffix: ``\_dcls''. We create a DCLS-empowered model by performing a drop-in replacement of all the depthwise separable convolutions of this model \citep{chollet2017xception,sandler2018mobilenetv2} with DCLS ones.

DCLS was introduced in \cite{hassani2023dilated}, where it exhibited better performance than the depthwise separable convolution and the dilated convolution \citep{yu2015multi} for computer vision tasks such as image classification, semantic segmentation, and object detection, as well as for computer audition tasks such as audio classification \citep{khalfaouiaudio}. Initially, the DCLS method used bilinear interpolation, in \cite{khalfaouihassani2023dilatedbeyond} this interpolation was extended to Gaussian. DCLS has focused on learning the positions of kernel elements along their weights. We believe this advance is important for tasks that require a nuanced understanding of visual context, similar to human perception.

%The ClickMe dataset provides a unique opportunity to assess how architectural enhancements like DCLS impact the interpretability of neural networks. Our work contributes to this discourse by investigating if and how the integration of DCLS into CNNs correlates with improved alignment to human visual strategies, thereby potentially increasing model transparency and interpretability.

%Bring the subject, from general to specific: why explainability in computer vision matters, what are the issues, what is your motivation, why DCLS in particular (because of its recent improvement in classification / other tasks segmentation object detect, we wanted to explore its explainability measure, scores ...)

\section{Methods}
\label{sec:methods}
%Talk about the experimental setup: datasets, metrics, scores ... 
\subsection{ClickMe dataset}

To quantitatively evaluate the interpretability of the models, we employed the ClickMe dataset \citep{linsley2019learning}, which was introduced to capture human attention strategies in classification tasks. The dataset collection process involved a single-player online game, ClickMe.ai \citep{linsley2019learning}, where players identified the most informative parts of an image for object recognition. The alignment of model-generated heatmaps with those from the ClickMe dataset measures how closely a model's attention strategy mirrors human strategy.

\subsection{DCLS method} 
%Dilated Convolution with Learnable Spacings (DCLS) is a technique introduced by \citet{hassani2023dilated} as an advancement in the field of convolutional neural networks (CNNs), particularly for enlarging the receptive field of the neural network without significantly increasing its number of parameters.

 Although larger convolution kernels can improve performance, increasing the kernel size increases the number of parameters and computational cost. \citet{yu2015multi} introduced dilated convolution (DC) to expand the kernel without increasing parameters. DC inserts zeros between kernel elements, effectively enlarging the kernel without adding new weights. However, DC uses a fixed grid, which can limit performance.

\citet{hassani2023dilated} presented DCLS as a new method that builds upon DC. Instead of using fixed spacings between non-zero elements in the kernel, DCLS allows learning these spacings through backpropagation. An interpolation technique is used to overcome the discrete nature of the spacings while maintaining the differentiability necessary for backpropagation.
%We aimed to determine if DCLS improves the model's interpretability, offering more insight into the decision-making process of the CNN.

\subsection{Grad-CAM and Threshold-Grad-CAM} 

Grad-CAM is a technique that provides visual explanations for the decisions made by deep neural networks. The method uses the gradients of a target concept, propagated into the final convolutional layer of a deep neural network, to produce a localization map highlighting the regions in the input image that are crucial for predicting this concept \citep{selvaraju2017grad}. Grad-CAM adapts to various network architectures by focusing on the last layer of interest before a classification head or pooling operation. The method is detailed in the supplementary material.

\subsubsection{Threshold-Grad-CAM} 

In the standard implementation of Grad-CAM, a ReLU activation is applied to the weighted combination of activation maps post-summation. This is predicated on the assumption that positive features should be exclusively highlighted as they are the ones contributing to the class prediction \citep{selvaraju2017grad}. However, our observations suggest that applying ReLU after the summation can inadvertently suppress useful signals when negative activations are present, as they may negate some positive activations when summed. This phenomenon becomes particularly pronounced in architectures such as ConvFormer and CAFormer, where we observed that the resulting heatmaps were no more informative than random heatmaps. We believe this is due to the choice of a specific activation: StarReLU in these two architectures \cite{yu2023metaformer}, which depends on two learnable parameters: scale and bias.
   
To address this issue, we propose applying ReLU to the activation maps before their summation. We then normalize the heatmaps. Finally, the heatmaps are thresholded to retain values above a predetermined threshold (determined experimentally to be $0.3$ for optimal results on the ClickMe dataset). The modified Grad-CAM process is described in the supplementary material. Our experiments demonstrated that this modification significantly improved the interpretability of the heatmaps generated for ConvFormer and CAFormer.

\section{Related work}
\label{sec:related}

The field of interpretable and explainable AI has recently gained significant attention within the AI community. Extensive research efforts range from defining key terms such as interpretability and explainability to developing explainability methods assessing their trustworthiness and evaluating the interpretability of deep learning models.
\cite{8631448} distinguished between interpretability and explainability and highlighted the challenge of achieving complete and interpretable explanations at the same time. \cite{1517589} defined interpretability as the ability to present model decisions in terms understandable to humans. In their study, \cite{10.1145/3397481.3450689} utilized multi-layer human attention masks to benchmark the effectiveness of explanation methods such as Grad-CAM and LIME. \cite{velmurugan2020evaluating} proposed functionally grounded evaluation metrics that assess the trustworthiness of explainability methods, including LIME and SHAP.
Furthermore, \cite{fel2022harmonizing} employed the ClickMe dataset to investigate the alignment between human and deep neural network (DNN) visual strategies, applying a training routine that aligns these strategies, as a result enhancing categorization accuracy.
In our study, we align with the interpretability definitions in the literature. We employ human heatmaps from the ClickMe dataset as ground truth to evaluate our model's interpretability.

\section{Experiments}
In this section, we present the experimental setup used to compare the performance and interpretability of our proposed models. Specifically, we calculated the top-1 accuracy of the models trained on the ImageNet1k validation dataset \citep{deng2009imagenet} to assess their classification effectiveness. For interpretability, we employed Spearman’s correlation as a metric to compare the alignment between human-generated heatmaps from the ClickMe dataset and the model-generated heatmaps. We assessed the interpretability of the heatmaps produced using two different methods: Grad-CAM and our proposed Threshold-Grad-CAM.
\subsection{Results}
\label{sec:results}

We present the results of integrating DCLS into state-of-the-art neural network architectures and our novel update to the Grad-CAM technique. Our experiments evaluated model interpretability, which we defined as the degree of alignment between heatmaps generated by explainability methods and those derived from human visualization strategies.

\subsection{Improvement in Model Interpretability with DCLS}

Our experiments incorporated DCLS into five model architectures: ResNet, ConvNeXt, CAFormer, ConvFormer, and FastViT. We trained each model on ImageNet1k. When this is mentioned by \_dcls, it means that the training has been done by replacing each depth-separable convolution of the baseline model with DCLS. 

\begin{figure}[!htbp]
  \centering
 % \fbox{
    \includegraphics[width=1\linewidth]{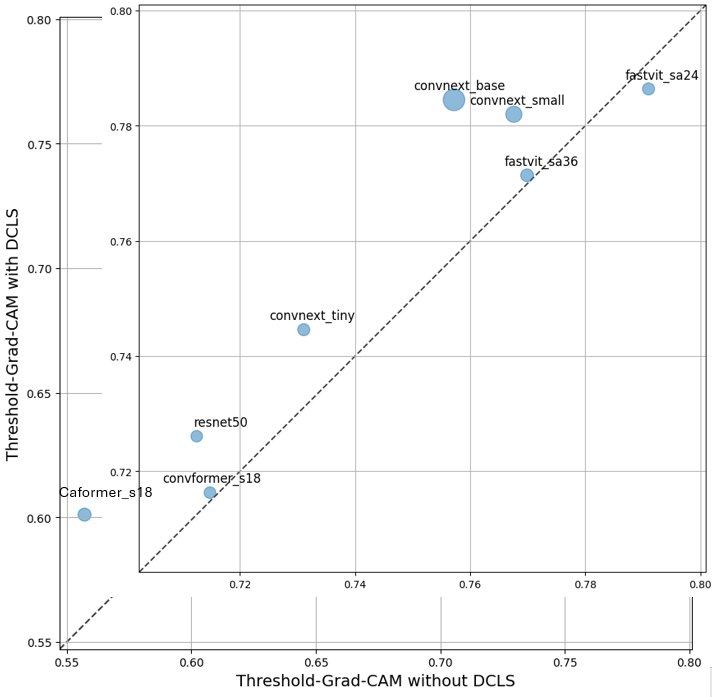} % Ensure this width fits within your document's layout
  %}
    \caption{Comparison of models interpretability score using Threshold-GradCAM with and without DCLS. Each point represents a different model, plotted according to its interpretability score without DCLS on the x-axis and with DCLS on the y-axis. Models above the dashed line demonstrate improved performance with the inclusion of DCLS.}
  \label{fig:comparison}
\end{figure}

The results showed an enhancement in model interpretability with all models but FastViT\_sa24. When equipped with DCLS, ConvNeXt improved in heatmap alignment. The score improved with both Grad-CAM and Threshold-Grad-CAM methods, as shown in Table~\ref{tab:dcls} and in Figure~\ref{fig:comparison}.
\begin{table}[!htbp]
    \centering
    \caption{Interpretability scores of various models on the ClickMe dataset using GradCAM and Threshold-GradCAM, with and without DCLS. The table presents the top-1 accuracy of each model alongside their respective interpretability scores. Models with the ``\textunderscore dcls" suffix indicate the use of DCLS.}
    \label{tab:dcls}
    \resizebox{0.5\textwidth}{!}{ 
    \begin{tabular}{lccc}
        %\hline
        Model & $\begin{array}{c}
        \text {Top1-} \\
        \text {accuracy}
        \end{array}$  & $\begin{array}{c}
        \text {Grad-CAM} \\
        \text {score}
        \end{array}$  & $\begin{array}{c}
        \text {Threshold-} \\
        \text {Grad-CAM score}
        \end{array}$  \\
        \hline
        convnext\_tiny & 82.1 & 0.6696 & 0.7311 \\
        \rowcolor{lightcream} convnext\_tiny\_dcls & 82.48 & 0.7561 & 0.7446 \\
        \hline
        convnext\_small & 83.15 & 0.7417 & 0.7676 \\
        \rowcolor{lightcream} convnext\_small\_dcls & 83.72 & 0.788 & 0.782 \\
        \hline
        convnext\_base & 83.83 & 0.7565 & 0.7572 \\
        \rowcolor{lightcream} convnext\_base\_dcls & 84.09 & 0.7979 & 0.7845 \\
        \hline
        fastvit\_sa24 & 81.11 & 0.7608 & 0.7933 \\
        \rowcolor{lightcream} fastvit\_sa24\_dcls & 82.48 & 0.7511 & 0.7754 \\
        \hline  
        fastvit\_sa36 & 82.85 & 0.6699 & 0.7699 \\
        \rowcolor{lightcream} fastvit\_sa36\_dcls & 82.69 & 0.699 & 0.7714 \\
        \hline         
        caformer\_s18 & 83.66 & 0.1719 & 0.5571 \\
        \rowcolor{lightcream} caformer\_s18\_dcls & 83.56 & -0.0594 & 0.6011 \\
        \hline
        convformer\_s18 & 82.98 & -0.0375 & 0.7148 \\
        \rowcolor{lightcream} convformer\_s18\_dcls & 83.06 & -0.1285 & 0.7163 \\
        \hline
        resnet50       & 77.84 & 0.6135 & 0.7125 \\
        \rowcolor{lightcream} resnet50\_dcls & 78.35 & 0.6252 & 0.7261 \\
        %\hline
    \end{tabular}
    }
\end{table}

Since Grad-CAM generates random heatmaps on CAFormer and ConvFormer architectures, we experimented with Threshold-Grad-CAM. Similar to ConvNeXt, CaFormer and ConvFormer showed higher interpretability scores when used with DCLS.
The FastViT\_sa24 model showed a high interpretability score, even without incorporating DCLS, and applying DCLS didn't improve the score.

In addition, DCLS increased the top-1 accuracy in all models but CAFormer\_s18 and FastViT\_sa36 (Table~\ref{tab:dcls}).

% \begin{table}[ht]
%     \centering
%     \caption{Write caption here}
%     \label{tab:baselines}
%     \resizebox{0.5\textwidth}{!}{ 

%     \begin{tabular}{|l|c|c|c|c|}
%         \hline
%         Model & \#parameters & Accuracy & Threshold-Grad-CAM \\
%         \hline
%         convnext\_tiny & 28M & 82.1  & 0.7311 \\
%         convnext\_small & 50M & 83.15  & 0.7676 \\
%         convnext\_base & 89M & 83.83  & 0.7572 \\
%         \hline
%         fastvit\_sa24 & 22M & 82.63  & 0.7563 \\
%         fastvit\_sa36 & 32M & 83.47 & 0.8011 \\
%         fastvit\_ma36 & 44M & 83.91& 0.8166 \\
%         \hline
%         hornet\_tiny\_7x7 & 22M & 82.83  & 0.7785 \\
%         hornet\_small\_7x7 & 50M & 83.79  & 0.79\\
%         hornet\_base\_7x7 & 87M & 84.23 & 0.7883 \\
%         \hline        
%         caformer\_s18 & 26M & 83.66  & 0.5571 \\
%         caformer\_s36 & 39M & 84.51 & 0.6794 \\
%         caformer\_m36 & 56M & 85.23 & 0.6858 \\
%         \hline
%         convformer\_s18 & 26M & 82.98  & 0.7148 \\
%         convformer\_s36 & 40M & 84.06 & 0.7264 \\
%         convformer\_m36 & 57M & 84.49 & 0.7357 \\
%         \hline
%     \end{tabular}
%     }
% \end{table}

%-------------------------------------------------------------------------

\section{Discussion}
\label{sec:discussion}
Except for FastViT, all model families studied show two points: first, an increase in accuracy when the depthwise separable convolution is replaced by DCLS, and second, an increase in the Treshold-Grad-CAM explanability score when this same modification is made. FastViT is a special model because the test inference is performed with a kernel reparametrization identical to that of RepLKNet \cite{ding2022scaling}. This could interfere with the DCLS method, which is in fact a different reparametrization, and might explain why the results for this family of models were not correlated in the same way as for the other studied models. Furthermore, the results presented here are significant since we tested three different training seeds for the ConvNeXt-T-dcls model and found an accuracy of $82.49 \pm 0.04$ and a Treshold-Grad-CAM score of $0.7466 \pm 0.004$.

\cite{fel2022harmonizing} utilized the ClickMe dataset to compare human and DNN visual strategies on ImageNet \citep{deng2009imagenet}. They adopted a classic explainability method: Image Feature Saliency \citep{jiang2015salicon}, to generate comparable feature importance maps for 84 deep neural networks (DNNs). They report that as DNNs become more accurate, a trade-off emerges where their alignment with human visual strategies starts to decrease.
In contrast, our study employs Grad-CAM for analyzing DNN visual strategies. Unlike the findings of \cite{fel2022harmonizing}, our use of Grad-CAM did not reveal such a trade-off. We think that this discrepancy is due to the differences in the explanatory methods used, which highlights the influence of analytical tools in interpreting DNN visual strategies.

Furthermore, it is conceivable that models with higher interpretability scores may focus more on those image features mostly correlated with the label class. A preliminary examination of the ClickMe dataset reveals that humans tend to concentrate solely on the object representing the class label within the image, ignoring other less directly related features to the class. This behavior likely stems from a nuanced human understanding of the concepts. Therefore, alignment with human-generated heatmaps might reflect a model's robustness.

\section{Conclusion}
In this study, we investigated the interpretability of recent deep neural networks using Grad-CAM-based methods for image classification tasks. We found that employing Dilated Convolution with Learnable Spacings enhances network interpretability. Our results indicate that DCLS-equipped models better align with human visual perception, suggesting that such models effectively capture conceptually relevant features akin to human understanding. Future work could focus on investigating the explainability score of DCLS using black-box methods such as RISE.

\bibliographystyle{named}
\bibliography{ijcai24}

\appendix
\onecolumn
\section{Appendix: Grad-CAM Implementation}

\begin{algorithm}
    \caption{Grad-CAM Implementation}
    \textbf{Input:} Image $I$, Target class $c$, Trained Convolutional Neural Network CNN\\
    \textbf{Output:} Heatmap $H$ visually highlighting influential regions for class $c$
    \begin{algorithmic}[1]
        \STATE \textbf{Forward Pass:} \\
        Process image $I$ through CNN to obtain feature maps at the last convolutional layer $A$. Let $A^k$ be the feature map for the $k$-th channel.
        
        \STATE \textbf{Compute Gradients:} \\
        Compute the gradient of the loss for class $c$, denoted $y^c$, with respect to the feature maps $A$, resulting in $\frac{\partial y^c}{\partial A^k}$.
        
        \STATE \textbf{Global Average Pooling of Gradients:} \\
        For each feature map channel $k$, compute the global average of the gradients:
        \[
        \alpha_k^c = \frac{1}{Z} \sum_i \sum_j \frac{\partial y^c}{\partial A_{ij}^k}
        \]
        where $i, j$ index spatial dimensions and $Z$ is the number of elements in $A^k$.
        
        \STATE \textbf{Weighted Combination of Feature Maps:} \\
        Compute the weighted sum of the feature maps using the weights $\alpha_k^c$:
        \[
        L^c = \text{ReLU}\left( \sum_k \alpha_k^c A^k \right)
        \]
        
        \STATE \textbf{Generate Heatmap:} \\
        Resize $L^c$ to the size of the input image $I$ to get the heatmap $H$.
        
        \STATE \textbf{Overlay Heatmap on Original Image:} \\
        Superimpose $H$ onto the original image $I$ for visualization, adjusting the transparency to ensure visibility of underlying features.
    \end{algorithmic}    
\end{algorithm}

\section{Appendix: Threshold-Grad-CAM Implementation}
\begin{algorithm}[!htbp]
    \caption{Threshold GradCAM}
    \textbf{Input:} Weighted activation maps $A^k$\\
    \textbf{Parameter:} Threshold value $t = 0.3$\\
    \textbf{Output:} Final heatmap $H$
    \begin{algorithmic}[1]
        \STATE \textbf{Apply ReLU Activation:} \\
        Apply the ReLU function to each weighted activation map to filter out negative values. This step prevents the cancellation of positive activations during summation:
        \[
        A^k_{\text{ReLU}} = \text{ReLU}\left( \alpha_k^c A^k\right) 
        \]
        
        \STATE \textbf{Summation of Activated Maps:} \\
        Sum the ReLU-activated maps:
        $
        S = \sum_k A^k_{\text{ReLU}}
        $
    
        \STATE \textbf{Normalization:} \\
        Normalize the summed activation map $S$ to ensure values are scaled consistently:
        $$
        N = \frac{S}{\max(S)}
        $$
    
        \STATE \textbf{Apply Thresholding:} \\
        Apply a threshold of $t$ to reduce noise and enhance the focus on relevant regions:
        $$
        H = \begin{cases} 
        N & \text{if } N \geq t \\
        0 & \text{otherwise}
        \end{cases}
        $$
    \end{algorithmic}
\end{algorithm}

Our revised approach yields more coherent and focused visual explanations, as validated by quantitative assessments of the ClickMe dataset. 
\newpage
\section{Appendix: DCLS vs. Baseline: Interpretability Analysis with Grad-CAM and Threshold-Grad-CAM}
\begin{figure*}[!htbp]
  \centering
  \begin{subfigure}{0.68\linewidth}
%    \fbox{\rule{0pt}{2in} \rule{.9\linewidth}{0pt}}
    \includegraphics[width=\linewidth]{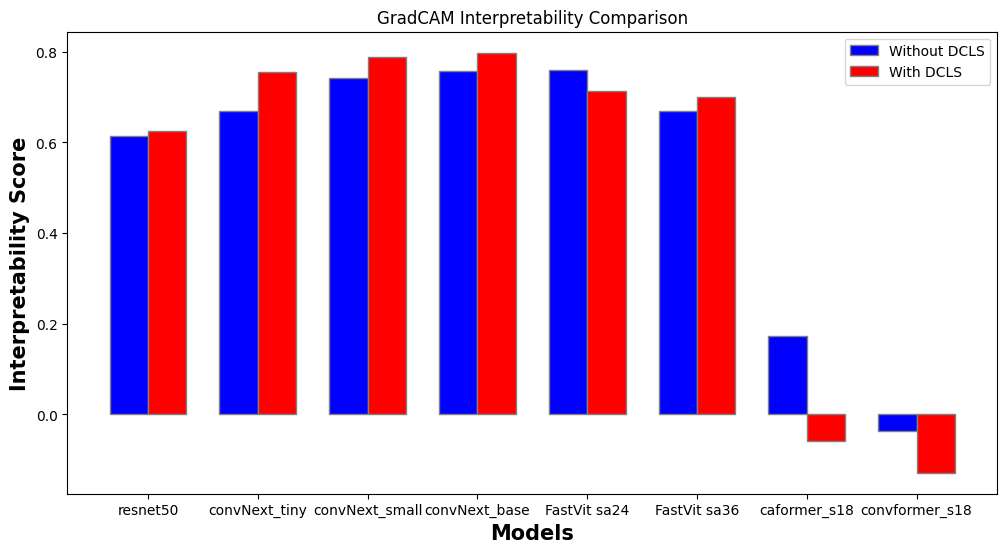}
    \label{fig:short-a}
  \end{subfigure}
  \hfill
  \begin{subfigure}{0.68\linewidth}
 %   \fbox{\rule{0pt}{2in} \rule{.9\linewidth}{0pt}}
    \includegraphics[width=\linewidth]{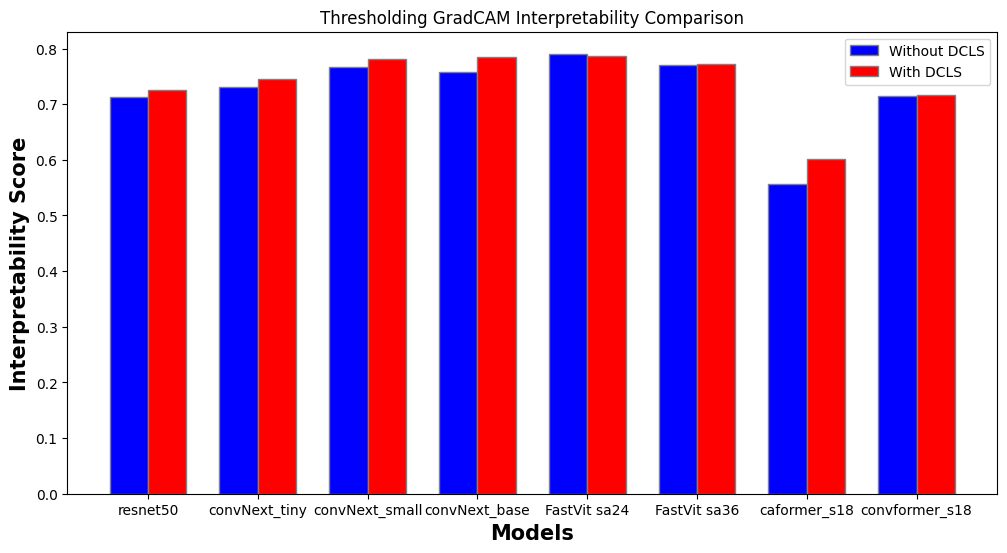}
    \label{fig:short-b}
  \end{subfigure}
  \caption{Comparative analysis of interpretability scores across different models using Grad-CAM and Threshold-Grad-CAM techniques. Top: The interpretability scores with Grad-CAM. Bottom: The interpretability scores with Threshold-Grad-CAM. Both subfigures highlight the difference in scores with and without DCLS. The results indicate that DCLS generally improves interpretability scores for most models.}
  \label{fig:short}
\end{figure*}
\newpage
\section{Appendix: Model Size vs. Interpretability Score Using Threshold-Grad-CAM}
\begin{figure*}[!htbp]
  \centering
 % \fbox{
    \includegraphics[width=\linewidth]{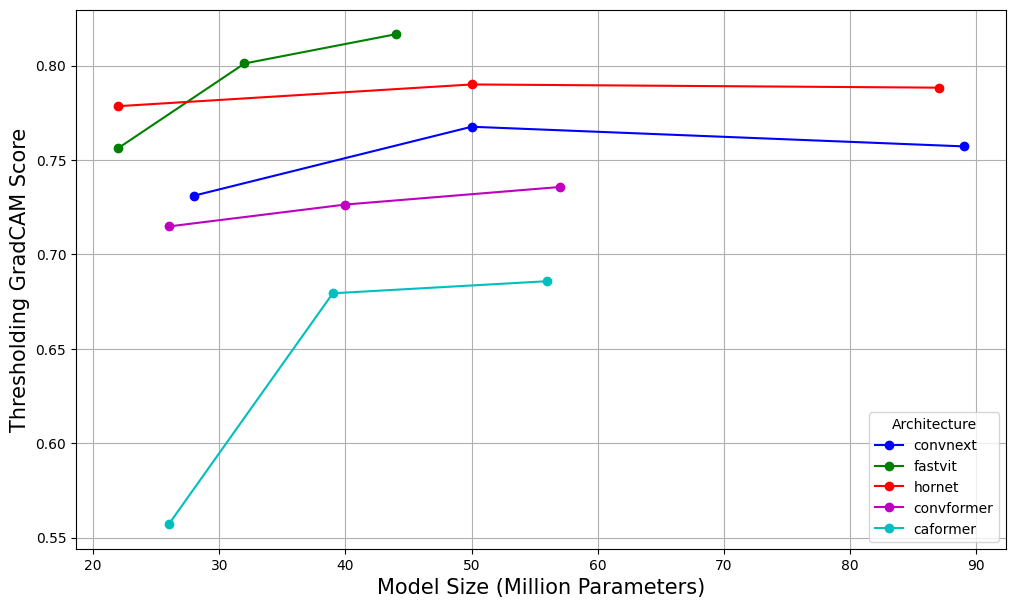} % Ensure this width fits within your document's layout
  %}
    \caption{Correlation between model size and Interpretability for baseline models, using Threshold-Grad-CAM scores. Larger models tend to have higher interpretability scores, suggesting a positive correlation between model size and explainability in baseline models.}
    \label{fig:baseline_model_size_interpretability}
\end{figure*}

\newpage
\section{Appendix: Visualizing Grad-CAM and Threshold-Grad-CAM Heatmaps}

\begin{figure*}[!htbp]
  \centering
    \includegraphics[width=\linewidth]{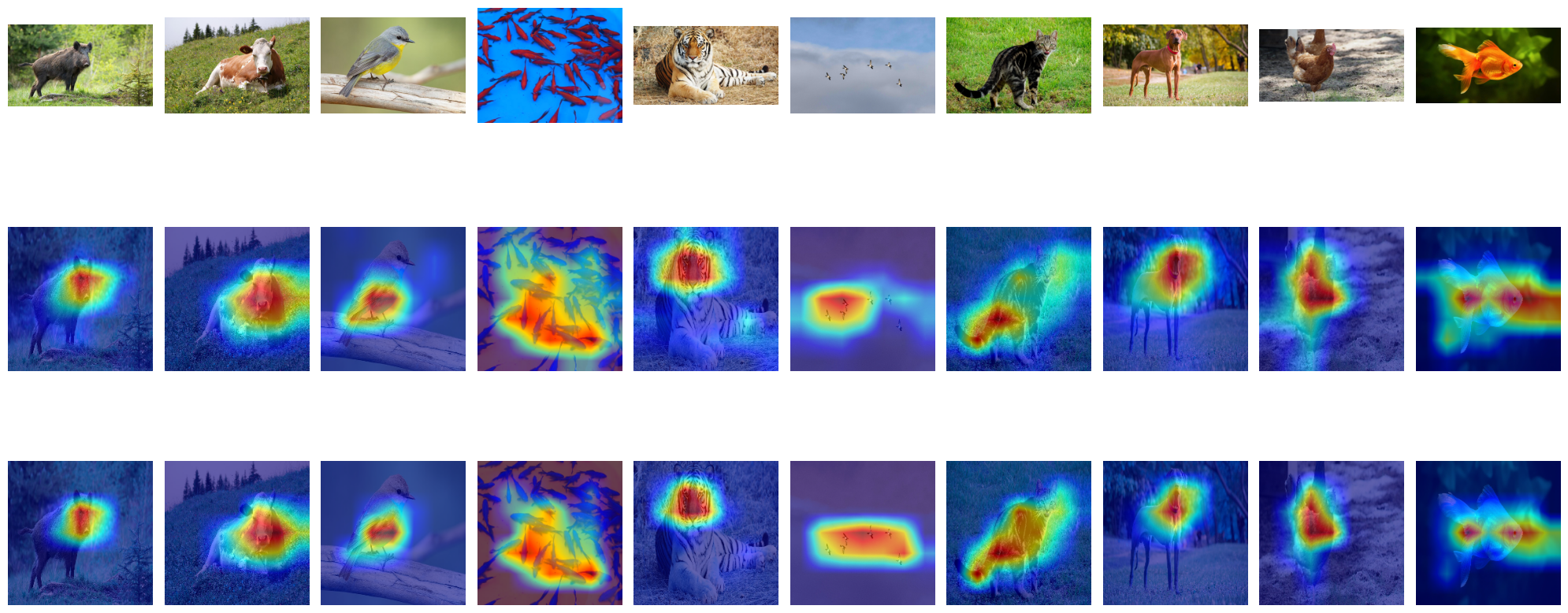}
    \caption{ResNet50 Grad-CAM heatmaps and Threshold-Grad-CAM heatmaps across 10 randomly chosen license-free internet images. Top row: Original images. Middle row: Images with Grad-CAM heatmaps. Bottom row: Images with Threshold-Grad-CAM heatmaps.}
    \label{fig:resnet_10images_comparison}
\end{figure*}

\begin{figure*}[!htbp]
  \centering
    \includegraphics[width=\linewidth]{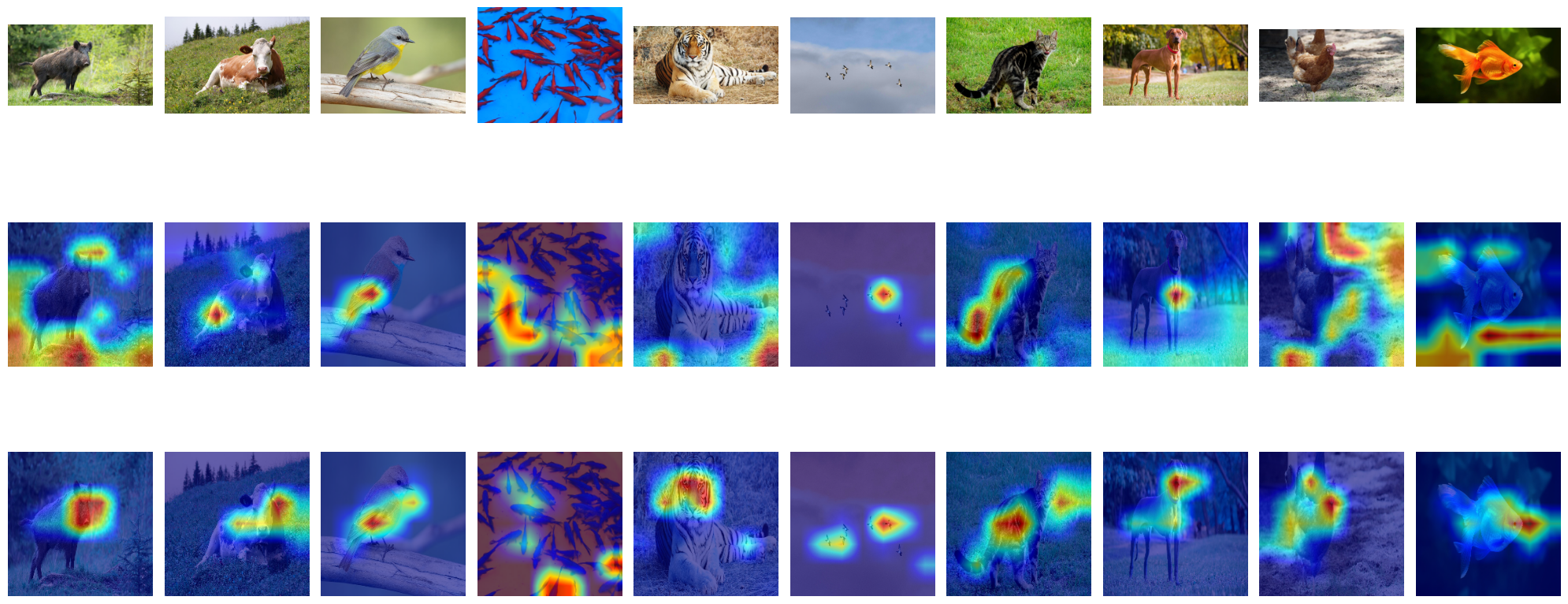}
    \caption{ConvFormer Grad-CAM heatmaps and Threshold-Grad-CAM heatmaps across 10 randomly chosen license-free internet images. Top row: Original images. Middle row: Images with Grad-CAM heatmaps. Bottom row: Images with Threshold-Grad-CAM heatmaps.}
    \label{fig:convformer_10images_comparison}
\end{figure*}

% \begin{figure}[t]
%   \centering
%     \includegraphics[width=\textwidth]{Interpretabityvssize.png}
%     \caption{Correlation between model size and interoperability for baseline models, using Thresholding GradCAM scores. The graph illustrates the interpretability score as a function of the number of parameters for each model. Each line corresponds to a different architecture, with the trend indicating that larger models tend to have higher interpretability scores, suggesting a positive correlation between model size and explainability in baseline models.}
%     \label{fig:baseline_model_size_interpretability}
% \end{figure}

% \begin{figure}[t]
%   \centering
%     \includegraphics[width=\textwidth]{Resnet_10images_comparison.png}
%     \caption{Resnet50 GradCAM heatmaps and Thresholding GradCAM heatmaps across 10 randomly chosen internet images. Top row: Original images. Middle row: Images with GradCAM heatmaps. Bottom row: Images with Thresholding GradCAM heatmaps.}
%     \label{fig:resnet_10images_comparison}
% \end{figure}

% \begin{figure}[t]
%   \centering
%     \includegraphics[width=\textwidth]{convformer.png}
%     \caption{Convformer GradCAM heatmaps and Thresholding GradCAM heatmaps across 10 randomly chosen internet images. Top row: Original images. Middle row: Images with GradCAM heatmaps. Bottom row: Images with Thresholding GradCAM heatmaps.}
%     \label{fig:convformer_10images_comparison}
% \end{figure}

\end{document}